\documentclass[runningheads]{llncs}
\usepackage[T1]{fontenc}
\usepackage{graphicx}
\begin{document}
\title{White Matter Geometry-Guided Score-Based Diffusion Model for Tissue Microstructure Imputation in Tractography Imaging}
\titlerunning{White Matter Geometry-Guided Diffusion Model for Imputation}
\author{Yui Lo\inst{1,2,4} \and Yuqian Chen \inst{1,4} \and Fan Zhang \inst{3} \and Dongnan Liu \inst{2} \and Leo Zekelman \inst{1,4}  \and Suheyla Cetin-Karayumak \inst{1,4} \and Yogesh Rathi\inst{1,4} \and Weidong Cai\inst{2} \and Lauren J. O'Donnell\inst{1,4}}
\authorrunning{Lo et al.}
\institute{Harvard Medical School, Boston, MA, USA\\
\email{odonnell@bwh.harvard.edu}
\and School of Computer Science, The University of Sydney, Sydney, NSW, Australia
\and University of Electronic Science and Technology of China, Chengdu, Sichuan, China
\and Brigham and Women’s Hospital, Boston, USA
}

\maketitle              % typeset the header of the contribution
\begin{abstract}
Parcellation of white matter tractography provides anatomical features for disease prediction, anatomical tract segmentation, surgical brain mapping, and non-imaging phenotype classifications. However, parcellation does not always reach 100\% accuracy due to various factors, including inter-individual anatomical variability and the quality of neuroimaging scan data. The failure to identify parcels causes a problem of missing microstructure data values, which is especially challenging for downstream tasks that analyze large brain datasets. In this work, we propose a novel deep-learning model to impute tissue microstructure: the White Matter Geometry-guided Diffusion (WMG-Diff) model. Specifically, we first propose a deep score-based guided diffusion model to impute tissue microstructure for diffusion magnetic resonance imaging (dMRI) tractography fiber clusters. Second, we propose a white matter atlas geometric relationship-guided denoising function to guide the reverse denoising process at the subject-specific level. Third, we train and evaluate our model on a large dataset with 9342 subjects. Comprehensive experiments for tissue microstructure imputation and a downstream non-imaging phenotype prediction task demonstrate that our proposed WMG-Diff outperforms the compared state-of-the-art methods in both error and accuracy metrics. Our code will be available at: https://github.com/SlicerDMRI/WMG-Diff.
\keywords{Diffusion MRI \and Score-based diffusion models \and Tractography \and Fractional anisotropy imputation \and Missing data}
\end{abstract}
\section{Introduction}
Diffusion magnetic resonance imaging (dMRI) tractography is the only non-invasive technique that can map the brain’s white matter (WM) connections and their tissue microstructure \cite{Zhang2022-eb}. Measures of tissue microstructure, such as fractional anisotropy (FA), are critical for studying the brain’s WM in health and disease across the lifespan \cite{Lebel2012-rg}. Microstructure quantification generally requires brain parcellation, with WM fiber clusters (FC) being a popular compact representation for statistical and machine-learning analyses of the WM \cite{Chen2023-ls,lo2024cross,Liu2023-lq,Zhang2018-vq,Xue2022-kh,Xue2023-bm}. Though fiber clustering is robust across datasets, acquisitions, and the human lifespan \cite{Zhang2019-do,Zhang2018-wh}, some fiber clusters can fail to be identified due to variability in scan quality across datasets \cite{Zhang2018-wh}, scan-specific factors such as subject motion art or scaifacts, inter-individual brain anatomical variability \cite{Thiebaut_de_Schotten2011-if}, or variability due to brain development and aging. The failure to identify fiber clusters causes a problem of missing microstructure data values, which is especially challenging for analyzing large brain datasets.

The problem of missing data in large brain datasets is a new topic receiving increasing attention. In the past year, novel methods have been proposed for imputing missing functional connectomes for individual subjects using predictive mean matching \cite{Liang2024-uq} and imputing missing morphometry values from an entire dataset using a graph neural network \cite{Wang2023-de}. These methods have improved performance on downstream tasks. However, we are unaware of prior work exploring imputation approaches dedicated to fiber cluster or dMRI microstructure data. This work investigates the imputation of FA values from an anatomically curated, atlas-based fiber cluster parcellation \cite{Zhang2018-wh} applied to a large, harmonized dMRI dataset \cite{Cetin-Karayumak2024-nd}.  

For imputation in this work, we employ score-based diffusion models, which have been successful in many deep generative modeling tasks \cite{Ho2020-fv,Song2021-ux,Tashiro2021-vo,Xiang2023-kf,zheng2022diffusion} and outperform competing state-of-the-art methods \cite{Ho2020-fv,Song2021-ux,Tashiro2021-vo,Xiang2023-kf,Yang2023-by,zheng2022diffusion}. The score-based diffusion model uses two Markov chains. Diffusion models show better distribution coverage, versatility, and stability compared to other generative models \cite{dai2023advdiff}. The forward process gradually introduces noise into the data across various time steps until the data is completely noisy. The score-based model learns to optimize through the output data distribution's gradient field (a.k.a., score) \cite{Song2021-ux,Yang2023-by}. The model learns to reconstruct the data to impute missing values during the reverse diffusion denoising process \cite{Ho2020-fv}. The conditional score-based diffusion model is an extension of the score-based diffusion model, where the conditional denoising function provides informative guidance toward the target \cite{Yang2023-by,zheng2022diffusion}. It is effective at learning the conditional probability distribution of the observable values in a given dataset to provide accurate imputation. The conditioning process has been shown to improve training speed and outperform the unconditional score-based model by utilizing the observable data’s scores \cite{Tashiro2021-vo,zheng2022diffusion}. During training, the conditional score-based model uses a random data selection process to create simulated observed and simulated missing data. Despite the effectiveness of diffusion models in various domains, we are unaware of their use in diffusion MRI microstructure analysis. This contribution marks an advance in leveraging diffusion models to analyze microstructural properties. 

In this work, we extend the score-based diffusion model. The main novelty and motivation of this work is to leverage information about WM fiber cluster geometry to define a guided denoising function. We propose a novel guided denoising function, which extends the conditional score-based diffusion model \cite{zheng2022diffusion} to replace a random data reverse diffusion process with a geometry-guided reverse diffusion process. In our guided approach to leverage prior knowledge about anatomical relationships, the approach allows the model to focus on fiber clusters geometrically similar to those actually missing in our dataset. This approach is motivated by the fact that geometrically similar fiber clusters generally represent similar anatomy with similar microstructural properties. By enabling training to focus on fiber clusters similar to missing clusters in each subject, we hypothesize this will guide the reverse diffusion process to impute more accurate tissue microstructure values.

We propose WMG-Diff, a White Matter Geometry-Guided Diffusion model, to impute missing dMRI microstructure data for a downstream prediction testbed task. The paper has three main contributions. First, our work introduces a deep generative approach using a conditional score-based diffusion model to impute the missing microstructure values. To our knowledge, this is the first work to explore dMRI microstructure imputation and apply generative diffusion models to this task. Second, we utilize the anatomical information of white matter geometry to create conditional masks that aid the reverse diffusion generation process. Specifically, we propose a guided denoising function informed by the geometric relationships between WM tractography fiber clusters, so that the function utilizes information from anatomically similar fiber clusters. Third, we evaluate our proposed method on a downstream testbed sex prediction task and show that WMG-Diff outperforms popular state-of-the-art imputation methods on a harmonized, large-scale dataset.

\section{Methods}
An overall conceptual representation of the framework used to impute missing FA values guided by geometric fiber cluster relationships is presented in Fig. 1. In this work, we focus on dMRI tractography data. We leverage a large dataset of fiber cluster microstructure measures. The dataset was created by generating whole brain tractography for each subject, parcellating subjects' WM fiber clusters, and extracting their quantitative FA measures \cite{Cetin-Karayumak2024-nd}. We identify the missing cluster FA values for each subject. Then, we calculate the geometric relationship between each pair of fiber clusters across the whole brain as a guided reference to the denoising function of our proposed WMG-Diff. The following paragraphs of this section describe the method in more detail.

\begin{figure}
\centering
\includegraphics[width=\textwidth]{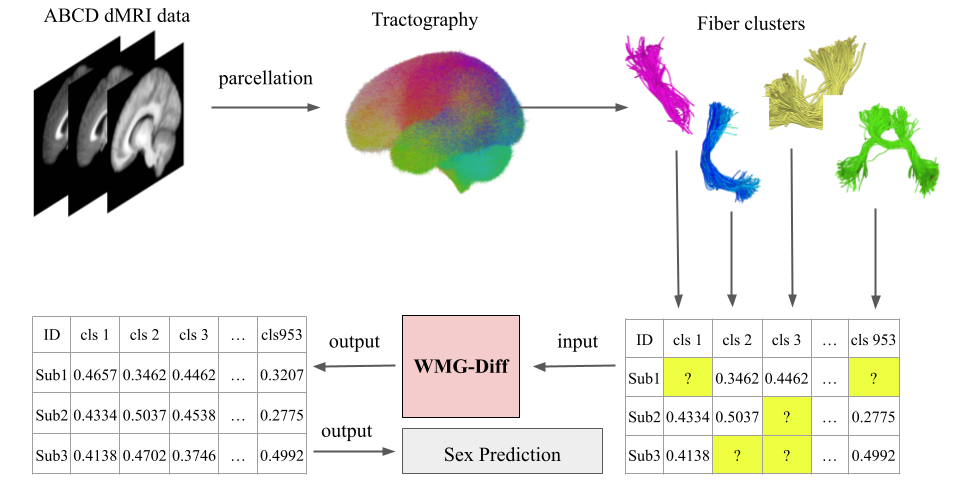}
\caption{Tractography parcellation into fiber clusters (top row) allows the calculation of FA values, represented in tabular format (bottom right), with missing values in yellow. The proposed WMG-Diff pipeline (bottom row) imputes missing FA data for input to a downstream sex prediction testbed task. Sub: Subject, cls: Cluster.} \label{fig1}
\end{figure}

\subsection{Dataset}
This study evaluates the model on the large-scale, multi-site Adolescent Brain Cognitive Development (ABCD) dataset, which includes 9,342 young children aged between 9 and 11 years \cite{Casey2018-fw,Cetin-Karayumak2024-nd}. To standardize the dMRI data across 21 different acquisition sites, we harmonized to remove scanner-specific biases while preserving inter-subject biological variability \cite{Cetin-Karayumak2024-nd}. The FC bundle parcellation employed was created to be consistent across subjects. This was demonstrated by the application of this FC parcellation across the human lifespan, across disease states, and across different acquisitions \cite{Zhang2018-wh}. The dataset includes 4,879 males (52.2\%) and 4,463 females (47.8\%). We divide the dataset into training (7,473 subjects, 80\%) and testing (1,869 subjects, 20\%) for five folds of model evaluation.

\subsection{Tractography and Fiber Clustering}
Tractography and white matter feature extraction was performed for the ABCD as follows \cite{Cetin-Karayumak2024-nd}. Whole brain tractography was generated for each individual subject’s dMRI data employing a two-tensor unscented Kalman filter method \cite{Malcolm2010-fj,Reddy2016-fv} to depict multiple crossing fibers. This method allows anatomically sensitive estimation of the pathway and connectivity of brain connections \cite{He2023-bl}. Tractography was subsequently parcellated into 953 fiber clusters using the ORG anatomically curated tractography white matter atlas \cite{Zhang2018-wh}. Each fiber cluster could contain hundreds of fibers, each illustrating a particular connection in the neural pathways. Microstructure measures including FA were computed for each streamline point from the multi-tensor model employed to perform tractography. The white matter cluster features were calculated as the mean FA value of all points within the cluster and input to the proposed framework.

\subsection{White Matter Atlas Geometric Distance}
We quantify the geometric relationship between the fiber clusters using a WM tractography distance, a common strategy widely used to cluster fibers \cite{Chen2021-jh,Chen2023-ls,Zhang2022-eb}. Fig. 2 illustrates three sample clusters from the ORG atlas \cite{Garyfallidis2012-ut,Zhang2018-wh}. The pairwise distance describes a cluster’s geometric relationship with the other clusters. We use the minimum average direct-flip pairwise streamline distance between each pair of fiber clusters in the ORG atlas \cite{Garyfallidis2012-ut,Zhang2018-wh}. Geometrically similar clusters demonstrate a low distance, indicating a high degree of similarity in white matter tractography anatomy, whereas a large distance suggests a lower level of similarity.

\begin{figure}
\centering
\includegraphics[width=\textwidth]{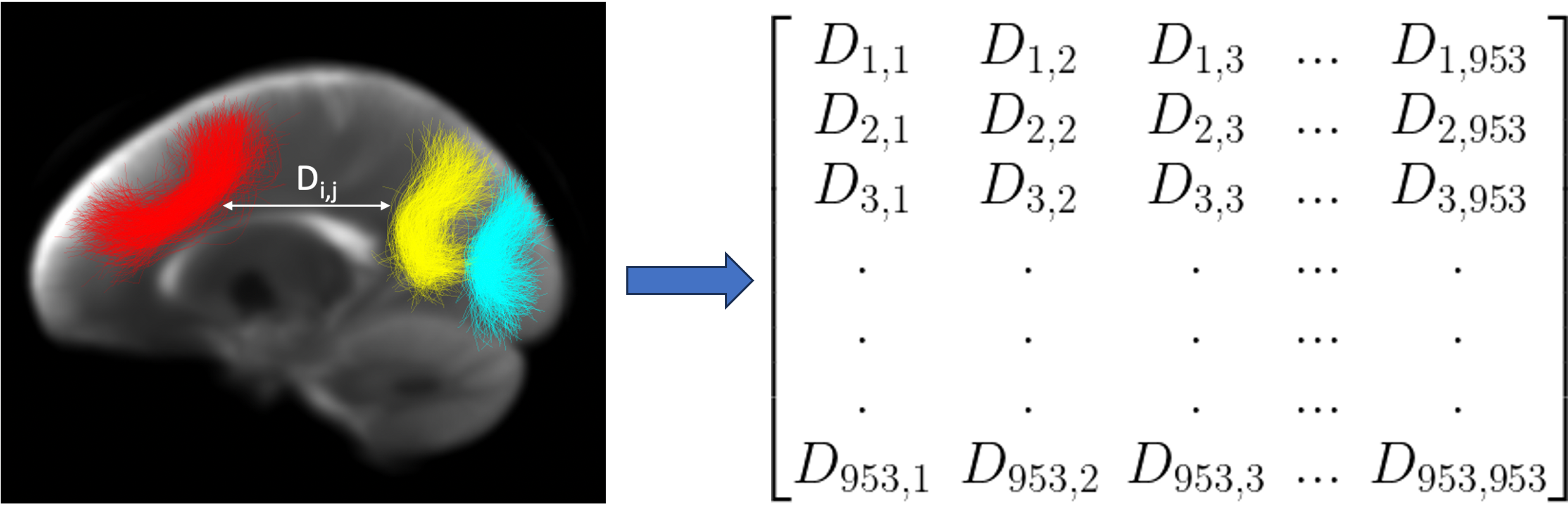}
\caption{An illustration of the pairwise distances (white arrows) between example fiber clusters (left) and the resulting matrix of distances between all of the 953 white matter fiber clusters (right). D: Distance.} \label{fig2}
\end{figure}

\subsection{Geometry-guided Score-based Diffusion Model}
To address missing FA data in tractography analysis, we propose to leverage a generative diffusion model. In particular, we adopt a score-based conditional diffusion model as it is effective at learning the conditional probability distribution of the observable values in a given dataset to provide accurate imputation \cite{zheng2022diffusion}. The model incorporates convolutional layers, a diffusion embedding module, transformer encoders, and residual blocks \cite{Tashiro2021-vo,zheng2022diffusion}. The training dataset comprises three parts: first, \textit{actual missing FA data}, which does not provide any ground-truth values for training. Then, the non-missing training data $x$ is randomly divided into the \textit{guided observable values} (80\%) and the \textit{imputation targets} (20\%) \cite{Tashiro2021-vo,zheng2022diffusion}. During self-supervised training, the imputation targets serve as synthetic missing data where the ground-truth data is known. In particular, the model’s residual block is conditioned with geometric guidance, which trains the denoising process through a continuous iterative diffusion process. The diffusion model estimates the true conditional data distribution of the imputation targets given the guided observable values distribution \cite{Tashiro2021-vo}.

For our application, our motivation for choosing the conditional score-based diffusion model is to impute the missing FA values, considering the subject-specific tissue microstructure distribution. We propose a geometric relationship-guided denoising function. Instead of a random selection of the non-missing training data, we leverage prior knowledge about the white matter structure to enhance the overall learning process about missing data. Our method aims to guide the training to focus on frequently missing FA data (e.g., particular fiber clusters often missing due to anatomical or scan protocol variability). We specifically select the guided observable values $x^{0}_{co}$ and the imputation targets $x^{0}_{ta}$ based on the geometric distances calculated from the white matter atlas in Section 2.2. In detail, for an individual training subject, we select the guided observable values from fiber clusters with minimal geometric distances to that subject’s actual missing fiber clusters. For each missing FA data value, K-guided observable values are selected (without repetition of data). K is chosen such that, in total, the guided observable values represent 80\% of the non-missing training data. This procedure leverages information about the WM anatomy and the distribution of missing fiber clusters to guide the training process.

\begin{figure}[htb]
\centering
\includegraphics[width=\textwidth]{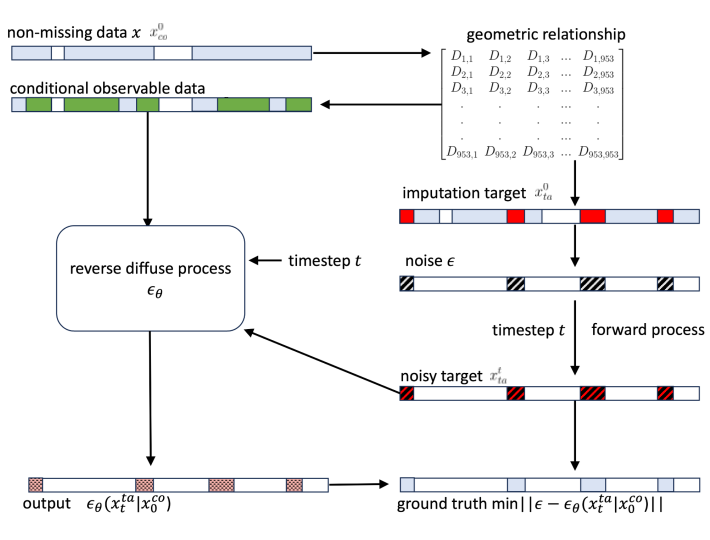}
\caption{Overview of the architecture of our WMG-Diff guided process. In the input (upper left), actual missing data (white) and non-missing observable data (blue) are shown. Then, the non-missing data is split into observable values (green) and imputation targets (red) using information about cluster geometric relationships. At timestep $t$, the imputation target undergoes the forward process to add noise until it becomes a fully noisy target. The reverse diffuse process leverages the guided observable data at each timestep to denoise the noisy target and minimize the difference between the noise and the output.} 
\label{fig3}
\end{figure}
\subsection{Downstream Sex Prediction Testbed Task}
To further evaluate the WMG-Diff model, we utilize the generated microstructure values from the model to serve as input for a downstream sex prediction task. The sex prediction task is a popular testbed task for evaluating the performance of deep learning methods \cite{chen2024tractgraphformer,Chen2023-ls,Wang2023-de}. Sex factors can provide insights into how cognitive functions develop and differ between males and females. We employ a machine learning logistic regression model, which has been widely applied to predict non-imaging phenotypes through neuroimaging features \cite{Chen2024-bu,Xue2022-rs,Cui2018-jc,Liu2023-tc,Zekelman2022-oo}. This downstream testbed task provides an additional metric (Accuracy) to understand the real-world applicability of the generated data where we define the downstream prediction as a binary problem.

\section{Experiments and Results}
\subsection{Implementation Detail}
The model includes four diffusion layers, each with 64 channels and two attention heads. The diffusion process is parameterized with an embedding dimension of 128, initiated with a beta of 0.0001 and ending beta of 0.5 across 150 steps, following a quadratic schedule for the backward diffusion process. The model is trained and evaluated with a batch size of 16 for 100 epochs. The learning rate is set to 0.0005. To define imputation targets, the missing ratio is selected to split the data into observable (80\%) and imputation targets (20\%) \cite{Tashiro2021-vo,zheng2022diffusion}. All experiments are performed on an NVIDIA RTX A5000 using PyTorch v1.13 \cite{Paszke2019-nb}.

To quantify the performance for the two tasks, the missing FA data imputation task and sex prediction testbed task, we employ the Root Mean Squared Error (RMSE) and the Accuracy (ACC) metrics, respectively. The RMSE metric is commonly used in generative models to evaluate the differences between the generated and actual data \cite{Tashiro2021-vo,Yoon2018-nv,zheng2022diffusion}. The ACC metric is commonly used to evaluate sex prediction in dMRI neuroimaging applications \cite{He2022-aq,Li2024-uo,Liu2023-tc}. ACC is defined as the percentage of correct predictions across all subjects.

\subsection{Comparison with SOTA Methods}
We conduct comparison experiments to determine the effectiveness of our proposed WMG-Diff. Unlike in medical image analysis, where interpolation of voxel data is the obvious baseline for imputation methods, there is no clear method to perform interpolation of FC microstructure data since it is tabular data and is not measured on a voxel grid. We compare common imputation methods in neuroimaging analysis that replace the missing data with the mean, median, or zero \cite{He2022-aq}. We then compare our approach to two open-source state-of-the-art imputation methods, Generative Adversarial Imputation Nets (GAIN) \cite{Yoon2018-nv} and Multivariate imputation by chained equations (MICE) \cite{Azur2011-xp}. GAIN is a method for imputing missing data by adapting the well-known Generative Adversarial Nets (GAN) framework. MICE is a regression method widely applied across different applications that has shown to be an efficient imputation strategy for downstream statistical analysis. Table 1 illustrates the performance of the compared methods. The results show that the WMG-Diff model outperforms state-of-the-art methods and common neuroimaging data imputation practices, indicated by the lowest RMSE value. 

\begin{table}
\centering
\caption{Comparison of imputation methods.}\label{tab1}
\begin{tabular}{|c|c|}
\hline
Imputation method & RMSE $\downarrow$\\
\hline
Mean & 0.1281 ± 0.0018\\
Median &  0.1275 ± 0.0014\\
Zero & 0.4213 ± 0.0002\\
GAIN \cite{Yoon2018-nv} &  0.0871 ± 0.0006 \\
MICE \cite{Azur2011-xp} &  0.0835 ± 0.0035\\
Baseline Diffusion \cite{zheng2022diffusion} &  0.0422 ± 0.0131\\
\textbf{Proposed} & \textbf{0.0310 ± 0.0036} \\
\hline
\end{tabular}
\end{table}
\subsection{Ablation Study}

We also evaluate the performance of our geometry-guided denoising function. We implement the conditional score-based diffusion model \cite{zheng2022diffusion} with randomly conditioned observable data and imputation targets as our baseline diffusion model method. In this experiment, we randomly drop 20\% of the FA values as synthetic missing data \cite{Tashiro2021-vo,Yoon2018-nv,zheng2022diffusion}, and we then impute with WMG-Diff and common imputation practices widely applied in neuroimaging applications of microstructure analysis. Results in Table 1 indicate that compared to using a random mask, a geometric relationship-guided denoising function achieves better performance.

\begin{table}
\centering
\caption{Downstream testbed sex prediction on imputed missing data.}\label{tab2}
\begin{tabular}{|c|c|}
\hline

Imputation method &  ACC $\uparrow$\\
\hline
Full input dataset &  0.7790 ± 0.0068\\
Mean  &  0.6993 ± 0.0247\\
Median &  0.6988 ± 0.0224\\
Zero & 0.5773 ± 0.0190\\
GAIN \cite{Yoon2018-nv} &  0.7373 ± 0.0291 \\
MICE \cite{Azur2011-xp}&  0.7330 ± 0.0165\\
Baseline Diffusion \cite{zheng2022diffusion}  &  0.7586 ± 0.0225\\
\textbf{Proposed} & \textbf{0.7712 ± 0.0078} \\
\hline
\end{tabular}
\end{table}

\subsection{Sex Prediction Testbed Evaluations}

To investigate the effectiveness of data imputation, we employ a logistic regression approach to predict sex based on the imputed FA values. Results in Table 2 show WMG-Diff outperforms widely applied imputation methods. We also compare the sex prediction performance using the full input dataset (without dropping 20\% as synthetic missing data). This comparison demonstrates that sex prediction accuracy using WMG-Diff imputation is almost identical to using the full dataset, showing the success of the proposed imputation method.

\section{Future Work}
In this study, we evaluated imputation methods on HCP data for gender prediction. Future evaluation of the effectiveness of WMG-Diff can be performed in a more comprehensive fashion by employing additional downstream tasks, such as age prediction \cite{Beheshti2022-ma}, in different datasets across age groups. The imputation of other tractography values, such as other measures of microstructure, shape, or connectivity  \cite{lo2024cross}, is of interest for future studies. It is also important to investigate additional methods, such as ensemble learning \cite{Ganaie2022-ed}, that may improve the overall imputation or the downstream prediction results. It is also important to take into consideration how generative AI models, such as WMG-Diff, utilize computation resources, in order to improve the overall computation efficiency.

\section{Conclusion}

In this paper, we proposed WMG-Diff, a conditional score-based diffusion-based model that utilizes the reverse diffusion process to impute missing FA values. We formulated the guided denoising function to leverage prior knowledge about the white matter structure geometric relationships between white matter fiber clusters to enhance learning during the reverse diffusion process. We evaluated the model on the ABCD dataset, and the results demonstrate that WMG-Diff outperformed widely applied SOTA methods. We then utilized the imputed data for a non-imaging phenotype prediction task. Results from WMG-Diff showed superior performance over widely applied imputation methods for the sex prediction testbed task. Overall, this suggests that the proposed WMG-Diff model can be used to generate accurate tissue microstructure data, which further benefits downstream tractography-related prediction tasks. 
\section{Compliance with Ethical Standards} 
This study uses public ABCD imaging data; Approval was granted by the BWH IRB for the use of the public. 
\section{Acknowledgement}
This work is supported by the following National Institutes of Health (NIH) grants: R01MH125860, R01MH132610, R01MH119222, R01NS125781, R01MH1-12748, R01AG042512, K24MH116366. This work is partly supported by the National Key R\&D Program of China (No. 2023YFE0118600) and the National Natural Science Foundation of China (No. 62371107). This work is also supported by The University of Sydney International Scholarship and Postgraduate Research Support Scheme.

\bibliographystyle{splncs04}
\bibliography{reference}

\begin{thebibliography}{10}
\providecommand{\url}[1]{\texttt{#1}}
\providecommand{\urlprefix}{URL }
\providecommand{\doi}[1]{https://doi.org/#1}

\bibitem{Azur2011-xp}
Azur, M.J., Stuart, E.A., Frangakis, C., Leaf, P.J.: Multiple imputation by chained equations: what is it and how does it work? International Journal of Methods in Psychiatric Research  \textbf{20}(1),  40--49 (2011)

\bibitem{Beheshti2022-ma}
Beheshti, I., Ganaie, M.A., Paliwal, V., Rastogi, A., Razzak, I., Tanveer, M.: Predicting brain age using machine learning algorithms: A comprehensive evaluation. IEEE J Biomed Health Inform  \textbf{26}(4),  1432--1440 (Apr 2022)

\bibitem{Casey2018-fw}
Casey, B.J., Cannonier, T., Conley, M.I., Cohen, A.O., Barch, D.M., Heitzeg, M.M., Soules, M.E., Teslovich, T., Dellarco, D.V., Garavan, H., Orr, C.A., Wager, T.D., Banich, M.T., Speer, N.K., Sutherland, M.T., Riedel, M.C., Dick, A.S., Bjork, J.M., Thomas, K.M., Chaarani, B., Mejia, M.H., Hagler, Jr, D.J., Daniela~Cornejo, M., Sicat, C.S., Harms, M.P., Dosenbach, N.U.F., Rosenberg, M., Earl, E., Bartsch, H., Watts, R., Polimeni, J.R., Kuperman, J.M., Fair, D.A., Dale, A.M., {ABCD Imaging Acquisition Workgroup}: The adolescent brain cognitive development ({ABCD}) study: Imaging acquisition across 21 sites. Developmental Cognitive Neuroscience  \textbf{32},  43--54 (Aug 2018)

\bibitem{Cetin-Karayumak2024-nd}
Cetin-Karayumak, S., Zhang, F., Zurrin, R., Billah, T., Zekelman, L., Makris, N., Pieper, S., O'Donnell, L.J., Rathi, Y.: Harmonized diffusion {MRI} data and white matter measures from the adolescent brain cognitive development study. Scientific Data  \textbf{11}(1), ~249 (2024)

\bibitem{Chen2024-bu}
Chen, Y., Zekelman, L.R., Zhang, C., Xue, T., Song, Y., Makris, N., Rathi, Y., Golby, A.J., Cai, W., Zhang, F., O'Donnell, L.J.: {TractGeoNet}: A geometric deep learning framework for pointwise analysis of tract microstructure to predict language assessment performance. Medical Image Analysis p. 103120 (Feb 2024)

\bibitem{Chen2021-jh}
Chen, Y., Zhang, C., Song, Y., Makris, N., Rathi, Y., Cai, W., Zhang, F., O'Donnell, L.J.: Deep fiber clustering: Anatomically informed unsupervised deep learning for fast and effective white matter parcellation. In: International Conference on Medical Image Computing and Computer Assisted Intervention (MICCAI) 2021. pp. 497--507. Springer International Publishing (2021)

\bibitem{chen2024tractgraphformer}
Chen, Y., Zhang, F., Wang, M., Zekelman, L.R., Cetin-Karayumak, S., Xue, T., Zhang, C., Song, Y., Makris, N., Rathi, Y., O'Donnell, L.J.: Tract{G}raph{F}ormer: Anatomically informed hybrid graph cnn-transformer network for classification from diffusion mri tractography. arXiv preprint arXiv:2407.08883  (2024)

\bibitem{Chen2023-ls}
Chen, Y., Zhang, F., Zekelman, L.R., Xue, T., Zhang, C., Song, Y., Makris, N., Rathi, Y., Cai, W., O'Donnell, L.J.: Tractgraphcnn: Anatomically informed graph {CNN} for classification using diffusion {MRI} tractography. In: 2023 {IEEE} 20th International Symposium on Biomedical Imaging ({ISBI}). pp.~1--5. IEEE (2023)

\bibitem{Cui2018-jc}
Cui, Z., Gong, G.: The effect of machine learning regression algorithms and sample size on individualized behavioral prediction with functional connectivity features. Neuroimage  \textbf{178},  622--637 (2018)

\bibitem{dai2023advdiff}
Dai, X., Liang, K., Xiao, B.: Advdiff: Generating unrestricted adversarial examples using diffusion models. arXiv preprint arXiv:2307.12499  (2023)

\bibitem{Ganaie2022-ed}
Ganaie, M.A., Hu, M., Malik, A.K., Tanveer, M., Suganthan, P.N.: Ensemble deep learning: A review. Eng. Appl. Artif. Intell.  \textbf{115}(105151),  105151 (Oct 2022)

\bibitem{Garyfallidis2012-ut}
Garyfallidis, E., Brett, M., Correia, M.M., Williams, G.B., Nimmo-Smith, I.: {QuickBundles}, a method for tractography simplification. Frontiers in Neuroscience  \textbf{6}, ~175 (2012)

\bibitem{He2022-aq}
He, H., Zhang, F., Pieper, S., Makris, N., Rathi, Y., Wells, W., O'Donnell, L.J.: Model and predict age and sex in healthy subjects using brain white matter features: A deep learning approach. In: 2022 {IEEE} 19th International Symposium on Biomedical Imaging ({ISBI}). pp.~1--5. IEEE (Mar 2022)

\bibitem{He2023-bl}
He, J., Zhang, F., Pan, Y., Feng, Y., Rushmore, J., Torio, E., Rathi, Y., Makris, N., Kikinis, R., Golby, A.J., O'Donnell, L.J.: Reconstructing the somatotopic organization of the corticospinal tract remains a challenge for modern tractography methods. Human Brain Mapping  \textbf{44}(17),  6055--6073 (Dec 2023)

\bibitem{Ho2020-fv}
Ho, J., Jain, A., Abbeel, P.: Denoising diffusion probabilistic models. Advances in Neural Information Processing Systems  \textbf{33},  6840--6851 (2020)

\bibitem{Lebel2012-rg}
Lebel, C., Gee, M., Camicioli, R., Wieler, M., Martin, W., Beaulieu, C.: Diffusion tensor imaging of white matter tract evolution over the lifespan. Neuroimage  \textbf{60}(1),  340--352 (2012)

\bibitem{Li2024-uo}
Li, S., Zhang, W., Yao, S., He, J., Zhu, C., Gao, J., Xue, T., Xie, G., Chen, Y., Torio, E.F., Feng, Y., Bastos, D.C., Rathi, Y., Makris, N., Kikinis, R., Bi, W.L., Golby, A.J., O'Donnell, L.J., Zhang, F.: Tractography-based automated identification of the retinogeniculate visual pathway with novel microstructure-informed supervised contrastive learning. bioRxiv  (Jan 2024)

\bibitem{Liang2024-uq}
Liang, Q., Jiang, R., Adkinson, B.D., Rosenblatt, M., Mehta, S., Foster, M.L., Dong, S., You, C., Negahban, S., Zhou, H.H., Chang, J., Scheinost, D.: Rescuing missing data in connectome-based predictive modeling. Imaging Neuroscience  \textbf{2},  1--16 (2024)

\bibitem{Liu2023-lq}
Liu, R., Li, M., Dunson, D.B.: {PPA}: Principal parcellation analysis for brain connectomes and multiple traits. Neuroimage  \textbf{276},  120214 (2023)

\bibitem{Liu2023-tc}
Liu, W., Chen, Y., Ye, C., Makris, N., Rathi, Y., Cai, W., Zhang, F., O'Donnell, L.J.: Fiber tract shape measures inform prediction of non-imaging phenotypes. arXiv preprint arXiv:2303.09124  (2023)

\bibitem{lo2024cross}
Lo, Y., Chen, Y., Liu, D., Liu, W., Zekelman, L., Zhang, F., Rathi, Y., Makris, N., Golby, A.J., Cai, W., O'Donell, L.J.: Cross-domain fiber cluster shape analysis for language performance cognitive score prediction. arXiv preprint arXiv:2403.19001  (2024)

\bibitem{Malcolm2010-fj}
Malcolm, J.G., Shenton, M.E., Rathi, Y.: Filtered multitensor tractography. IEEE Transactions on Medical Imaging  \textbf{29}(9),  1664--1675 (2010)

\bibitem{Paszke2019-nb}
Paszke, A., Gross, S., Massa, F., Lerer, A., Bradbury, J., Chanan, G., Killeen, T., Lin, Z., Gimelshein, N., Antiga, L., Desmaison, A., K{\"o}pf, A., Yang, E., DeVito, Z., Raison, M., Tejani, A., Chilamkurthy, S., Steiner, B., Fang, L., Bai, J., Chintala, S.: {PyTorch}: an imperative style, high-performance deep learning library. In: Proceedings of the 33rd International Conference on Neural Information Processing Systems, pp. 8026--8037. Curran Associates Inc., Red Hook, NY, USA (Dec 2019)

\bibitem{Reddy2016-fv}
Reddy, C.P., Rathi, Y.: Joint {Multi-Fiber} {NODDI} parameter estimation and tractography using the unscented information filter. Frontiers in Neuroscience  \textbf{10}, ~166 (2016)

\bibitem{Thiebaut_de_Schotten2011-if}
Thiebaut~de Schotten, M., Ffytche, D.H., Bizzi, A., Dell'Acqua, F., Allin, M., Walshe, M., Murray, R., Williams, S.C., Murphy, D.G.M., Catani, M.: Atlasing location, asymmetry and inter-subject variability of white matter tracts in the human brain with {MR} diffusion tractography. Neuroimage  \textbf{54}(1),  49--59 (Jan 2011)

\bibitem{Song2021-ux}
Song, Y., Durkan, C., Murray, I., Ermon, S.: Maximum likelihood training of score-based diffusion models. Advances in Neural Information Processing Systems  \textbf{34},  1415--1428 (2021)

\bibitem{Tashiro2021-vo}
Tashiro, Y., Song, J., Song, Y., Ermon, S.: {CSDI}: Conditional score-based diffusion models for probabilistic time series imputation. Advances in Neural Information Processing Systems  \textbf{34},  24804--24816 (2021)

\bibitem{Wang2023-de}
Wang, Y., Peng, W., Tapert, S.F., Zhao, Q., Pohl, K.M.: Imputing brain measurements across data sets via graph neural networks. In: Predictive Intelligence in Medicine. pp. 172--183. Springer Nature Switzerland (2023)

\bibitem{Xiang2023-kf}
Xiang, T., Yurt, M., Syed, A.B., Setsompop, K., Chaudhari, A.: {DDM}$^2$: Self-supervised diffusion mri denoising with generative diffusion models. In: The Eleventh International Conference on Learning Representations (2023)

\bibitem{Xue2022-rs}
Xue, T., Zhang, F., Zekelman, L.R., Zhang, C., Chen, Y., Cetin-Karayumak, S., Pieper, S., Wells, W.M., Rathi, Y., Makris, N., et~al.: Tractoscr: a novel supervised contrastive regression framework for prediction of neurocognitive measures using multi-site harmonized diffusion mri tractography. Frontiers in Neuroscience  \textbf{18} (2024)

\bibitem{Xue2023-bm}
Xue, T., Zhang, F., Zhang, C., Chen, Y., Song, Y., Golby, A.J., Makris, N., Rathi, Y., Cai, W., O'Donnell, L.J.: Superficial white matter analysis: An efficient point-cloud-based deep learning framework with supervised contrastive learning for consistent tractography parcellation across populations and {dMRI} acquisitions. Med. Image Anal.  \textbf{85}(102759),  102759 (Apr 2023)

\bibitem{Xue2022-kh}
Xue, T., Zhang, F., Zhang, C., Chen, Y., Song, Y., Makris, N., Rathi, Y., Cai, W., O'Donnell, L.J.: Supwma: Consistent and efficient tractography parcellation of superficial white matter with deep learning. In: 2022 IEEE 19th International Symposium on Biomedical Imaging (ISBI). pp.~1--5. IEEE (Mar 2022)

\bibitem{Yang2023-by}
Yang, L., Zhang, Z., Song, Y., Hong, S., Xu, R., Zhao, Y., Zhang, W., Cui, B., Yang, M.H.: Diffusion models: A comprehensive survey of methods and applications. ACM Computing Surveys  \textbf{56}(4),  1--39 (Nov 2023)

\bibitem{Yoon2018-nv}
Yoon, J., Jordon, J., van~der Schaar, M.: {{GAIN}}: Missing data imputation using generative adversarial nets. In: Dy, J., Krause, A. (eds.) Proceedings of the 35th International Conference on Machine Learning. Proceedings of Machine Learning Research, vol.~80, pp. 5689--5698. Proceedings of Machine Learning Research (2018)

\bibitem{Zekelman2022-oo}
Zekelman, L.R., Zhang, F., Makris, N., He, J., Chen, Y., Xue, T., Liera, D., Drane, D.L., Rathi, Y., Golby, A.J., O'Donnell, L.J.: White matter association tracts underlying language and theory of mind: An investigation of 809 brains from the human connectome project. Neuroimage  \textbf{246},  118739 (2022)

\bibitem{Zhang2022-eb}
Zhang, F., Daducci, A., He, Y., Schiavi, S., Seguin, C., Smith, R.E., Yeh, C.H., Zhao, T., O'Donnell, L.J.: Quantitative mapping of the brain's structural connectivity using diffusion {MRI} tractography: A review. Neuroimage  \textbf{249},  118870 (2022)

\bibitem{Zhang2018-vq}
Zhang, F., Wu, W., Ning, L., McAnulty, G., Waber, D., Gagoski, B., Sarill, K., Hamoda, H.M., Song, Y., Cai, W., Rathi, Y., O'Donnell, L.J.: Suprathreshold fiber cluster statistics: Leveraging white matter geometry to enhance tractography statistical analysis. Neuroimage  \textbf{171},  341--354 (May 2018)

\bibitem{Zhang2019-do}
Zhang, F., Wu, Y., Norton, I., Rathi, Y., Golby, A.J., O'Donnell, L.J.: Test-retest reproducibility of white matter parcellation using diffusion {MRI} tractography fiber clustering. Human Brain Mapping  \textbf{40}(10),  3041--3057 (2019)

\bibitem{Zhang2018-wh}
Zhang, F., Wu, Y., Norton, I., Rigolo, L., Rathi, Y., Makris, N., O'Donnell, L.J.: An anatomically curated fiber clustering white matter atlas for consistent white matter tract parcellation across the lifespan. Neuroimage  \textbf{179},  429--447 (2018)

\bibitem{zheng2022diffusion}
Zheng, S., Charoenphakdee, N.: Diffusion models for missing value imputation in tabular data. In: Conference on Neural Information Processing Systems (NeurIPS) Table Representation Learning (TRL) Workshop (2022)

\end{thebibliography}
%Author, F.: Article title. Journal \textbf{2}(5), 99--110 (2016)

%Author, F., Author, S.: Title of a proceedings paper. In: Editor,
%F., Editor, S. (eds.) CONFERENCE 2016, LNCS, vol. 9999, pp. 1--13.
%Springer, Heidelberg (2016). \doi{10.10007/1234567890}

%Author, F., Author, S., Author, T.: Book title. 2nd edn. Publisher,

%Author, A.-B.: Contribution title. In: 9th International Proceedings
%on Proceedings, pp. 1--2. Publisher, Location (2010)

%LNCS Homepage, \url{http://www.springer.com/lncs}, last accessed 2023/10/25
%\end{thebibliography}
\end{document}